\documentclass[10pt,twocolumn,letterpaper]{article}

\usepackage{cvpr}
\usepackage{times}
\usepackage{epsfig}
\usepackage{graphicx}
\usepackage{amsmath}
\usepackage{amssymb}

\usepackage{algorithm}
\usepackage[noend]{algpseudocode}
\usepackage{epstopdf}

\makeatletter
\def\BState{\State\hskip-\ALG@thistlm}
\makeatother

\cvprfinalcopy 

\usepackage[pagebackref=true,breaklinks=true,letterpaper=true,colorlinks,bookmarks=false]{hyperref}

\def\cvprPaperID{4} 

\ifcvprfinal\fi

\begin{document}

\title{HadaNets: Flexible Quantization Strategies for Neural Networks}

\author{Yash Akhauri\\
Birla Institute of Technology and Science, Pilani\\
India, RJ 333031\\
{\tt\small f2016142@pilani.bits-pilani.ac.in}
}
 
\maketitle

\begin{abstract}
    On-board processing elements on UAVs are currently inadequate for training and inference of Deep Neural Networks. This is largely due to the energy consumption of memory accesses in such a network. HadaNets introduce a flexible train-from-scratch tensor quantization scheme by pairing a full precision tensor to a binary tensor in the form of a Hadamard product. Unlike wider reduced precision neural network models, we preserve the train-time parameter count, thus out-performing XNOR-Nets without a train-time memory penalty. Such training routines could see great utility in semi-supervised online learning tasks. Our method also offers advantages in model compression, as we reduce the model size of ResNet-18 by $7.43\times$ with respect to a full precision model without utilizing any other compression techniques. We also demonstrate a 'Hadamard Binary Matrix Multiply' kernel, which delivers a 10-fold increase in performance over full precision matrix multiplication with a similarly optimized kernel. 
\end{abstract}

\section{Introduction}

Convolutional neural networks (CNNs) have achieved state-of-the-art results on several computer vision tasks~\cite{behemothHe_2016_CVPR, VGGnet} and even demonstrate superhuman classification in some test-cases~\cite{Superhuman}. Such models are often the result of ensemble learning~\cite{Ensemble} and behemoth neural networks~\cite{behemothHe_2016_CVPR}. At the other end, there is a huge demand for offline inference on embedded devices. Aligning with progress in this field, low power inference and training is an intriguing domain, drawing attention from corporations and researchers. 

State-of-the-art Convolutional Neural Networks typically require large amounts of memory and computation. This is not entirely feasible for edge devices, with ResNet-152~\cite{behemothHe_2016_CVPR} holding 60M parameters and 11 GFLOPS for a single forward pass on a single element batch. It is unrealistic to use such classifiers on current mobile devices.

Current attempts at reducing the memory footprint of neural networks focus primarily on compression techniques~\cite{DeepCompress, Compress3, Compress2, Compress1} or bit-width reduction~\cite{CourbariauxB1, Gupta15, Das18, LowPandSparisity, TTertQuantization, TWN}. Several of these techniques are not completely realizable on current hardware. 

XNOR-Nets~\cite{RastegariX1} have an exceptionally low memory footprint and energy requirement. Most multiply-accumulate operations are replaced by simple bit-manipulations~\cite{CourbariauxB1}. Aside from sequential bit-packing for subsequent bit-operators (XNOR operation), binary neural networks are fully realizable on current hardware. Xcel-RAM~\cite{Agrawal2018XcelRAMAB} is a modified von-Neumann machine which enables binary convolutions, providing a $6.1\times$ energy saving for XNOR-Net inference. The scope for such hardware optimization further incentivizes the search for better quantization strategies for binary neural networks.\\
This paper makes the following contributions:
\begin{itemize}
\item[$\bullet$] We introduce a quantized neural network training strategy with flexible memory and energy requirements. HadaNets yield a higher accuracy than XNOR-Nets without increasing filter map counts, as opposed to WRPN Nets~\cite{Mishra17}. 
\item[$\bullet$] HadaNets prove the be a more effective training strategy for quantized neural networks as both XNOR-Nets and HadaNets require us to maintain full-precision parameters for gradient updates. Our training methodology performs at par with post-training quantization methodologies like ABC-Nets~\cite{LinABC} ($\pm 0.5\%$ top-1 accuracy). 
\item[$\bullet$] We introduce Hadamard-Binary-Weight-Networks (HBWNs) (indicated by $\beta_{a} = 1$ in our tests). HBWNs outperform Binary-Weight-Networks~\cite{RastegariX1} on the ResNet-18 topology by 1.5\% in top-1 accuracy.
\item[$\bullet$] We develop Hadamard binary matrix multiply CPU kernel which demonstrates a 10 fold increase in performance over its full precision counter-part. 

\end{itemize}

\section{Related Work}
Aiming to reduce the computation and memory costs of current architectures, bit quantization and compression techniques have been thoroughly explored in recent literature. We briefly review these works in this section.
\subsection{Quantizing pre-trained models}
DeepCompression~\cite{DeepCompress} prunes unimportant connections, after which the remaining weights are quantized, followed by Huffman coding to compress the weight values. Incremental Network Quantization~\cite{INQZhou} involves weight partitioning, followed by group-wise quantization and re-training. ABC Networks~\cite{LinABC} deliver a performance identical to its full precision counterparts given sufficient activation and weight bases. Methods such as Network Sketching~\cite{NetSketch} use full precision activations for forward passes, they do not exploit the massive memory and energy saving that is offered by bit-packing binary matrix elements.

\subsection{Train-from-scratch quantization}
The current standard (32 bit weights and activations) has proven to be unnecessary for high accuracy in computer vision.~\cite{Gupta15} and~\cite{Das18} demonstrate that 16 bit precision is sufficient to train most neural networks. Binary Neural Networks~\cite{CourbariauxB1} have been proven trainable for small datasets, with XNOR-Nets~\cite{RastegariX1} converging for the ImageNet dataset. Mishra \etal~\cite{Mishra17} demonstrate that the accuracy loss by binarization can be prevented by increasing the filter map count in each layer, this increases the parameter count quadratically~\cite{Banner18}. This raises questions about the efficiency of this approach. Our research differs from these as we do not increase the train-time parameter count or the width of the convolutional filters, thus making our method less resource intensive to train.

\begin{figure*}
\begin{center}
    \includegraphics[width=1\linewidth]{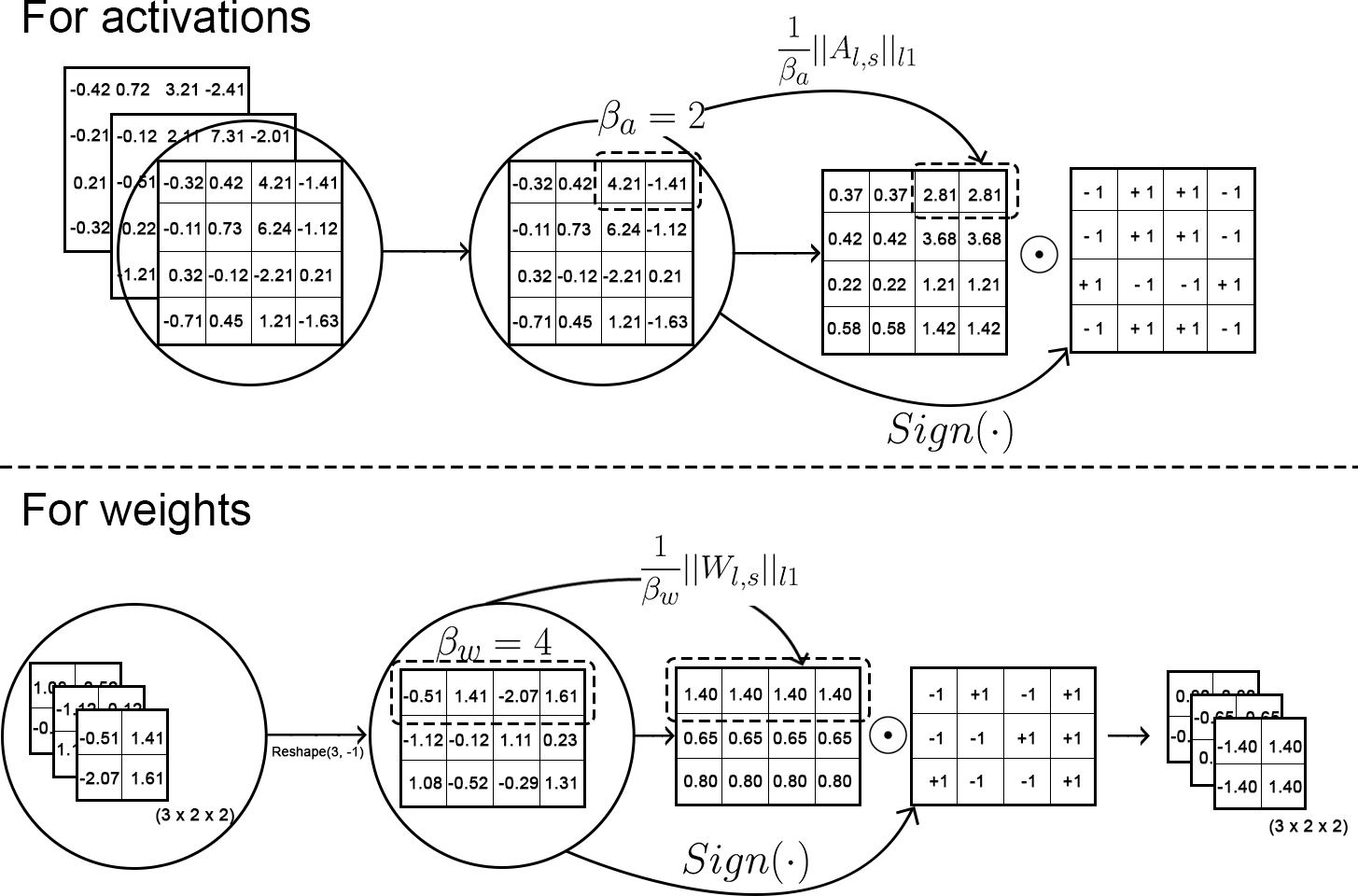}
\end{center}
   \caption{The figure depicts the 'hadamard' binarization of tensors in the CNN. We do not store repeated full precision values and compress the binary matrix by bit-packing it to available data-types. $A_{l, s}$ and $W_{l, s}$ refer to sub-matrices of size $(\beta \times 1)$ extracted from the full precision matrix of layer 'l' from an L-layer CNN. In our figure, we only show the Hadamard Binarization of one channel for the activations.}
\label{hadavisual}
\end{figure*}
\section{HadaNets}
HadaNets extend the binarization scheme used in XNOR-Nets~\cite{RastegariX1} by converting an input tensor to the hadamard product of a full precision tensor and a binary tensor. The binary tensor is simply the $Sign$ of the input tensor. The full precision tensor holds the mean of the absolute value of segments of the input tensor. The segment size is decided by $\beta$. This is depicted in Figure \ref{hadavisual}. \\
We refer to the input tensor at the $l^{th}$ layer in an L-layer CNN architecture with $A_{l (l=1,..,L)}$. We refer to the $k^{th}$ weight filter in the $l^{th}$ layer as $W_{l k (k=1,...,K^{l})}$. $K^{l}$ is the number of weight filters in the $l^{th}$ layer of the CNN.
$A_{l} \in \mathbb{R}^{c \times w_{in} \times h_{in}}$ where $(c, w_{in}, h_{in})$ represents \textit{channels, width} and \textit{height} respectively. $W_{l} \in \mathbb{R}^{k \times c \times w \times h}$ $s.t.$  $w \leq w_{in}, h \leq h_{in}$. The 'hadamard binarized' tensors as calculated by Eqn \eqref{eq2} and Eqn \eqref{eq4} in Section 3.1.1 and 3.1.2 for $W_{l}$ and $A_{l}$ respectively will be referenced as $\widetilde{W_{l}}$ and $\widetilde{A_{l}}$.

\begin{algorithm}
\caption{Training a L-Layer HadaNet}\label{euclid}
 \hspace*{\algorithmicindent} \textbf{Input:} (Input, Target): $(I,T)$, Cost function: $C(T, \widehat{T})$, Current LR: $\eta^{t}$, set $\beta_{a,l}$ and $\beta_{w,l}$.\\
 \hspace*{\algorithmicindent} \textbf{Output:} Updated weight $W^{t+1}$ and learning rate $\eta^{t+1}$
\begin{algorithmic}[1]
\For{$l$ = 1 to L}
    \State $S_{l}$ = \textbf{Shape(}$W_{l}$\textbf{)}
    \State $X_{l}$ = \textbf{Reshape(}$W_{l}$, shape=$(S_{l}[0], -1)$\textbf{)} // \textit{Reshape with -1 infers the second dimension from the length of the array and remaining dimensions.}
    \State $\widetilde{X_{l}}$ = \textbf{Binarize(}$X_{l}, \beta_{w}$\textbf{)} // \textit{Binarize as Eqn \eqref{eq2}}
    \State $\widetilde{W_{l}}$ = \textbf{Reshape(}$X_{l}$, shape= $S_{l}$\textbf{)}
\EndFor\\
$\widehat{T}$ = \textbf{HadaForward}($I$, $\widetilde{W}$) // \textit{Standard forward propagation with activations binarized as Eqn \eqref{eq4}}   \\
$\frac{\delta C}{\delta W}$ = \textbf{HadaBackward}($\frac{\delta C}{\delta \widetilde{W}}$, $\widetilde{W}$) // \textit{Standard backward propagation, gradients are calculated using $\widetilde{W}$}.   \\
$W^{t+1}$ = \textbf{UpdateParameters}($W^{t}$, $\frac{\delta C}{\delta W}$, $\eta_{t}$)  \\
$\eta_{t+1}$ = \textbf{UpdateLearningRate}($\eta^{t}$, t)
\end{algorithmic}
\label{algorithm1}
\end{algorithm}
\subsection{Hadamard binarization of tensors}
We introduce two hyper-parameters per layer in a neural network. We refer to these as $\beta_{w, l}$ and $\beta_{a, l}$ where $l$ is the $l^{th}$ layer in a L-Layer CNN and suggest $\beta_{w, l \in \{0, L\}} = \beta_{a, l \in \{0, L\}} = 1$ for standard architectures. Figure \ref{hadavisual} depicts the process of binarization and the role of $\beta_{a,l}$ and $\beta_{w,l}$. $\beta$ is also referred to as the binarization aggression in this paper. In all our experiments, $\beta_{w, l \in \{0, L\}} = \beta_{a, l \in \{0, L\}} = 1$. If $'l'$ is not sub-scripted for $\beta_{w}$ or $\beta_{a}$, we take $\beta_{w,l\in[1,L-1]} = \beta_{w}$ and  $\beta_{a, l\in[1,L-1]} = \beta_{a}$ in a L-Layer CNN.\\
$\odot$ denotes Hadamard product of two tensors. \\$\delta \in \mathbb{R}^{layer \times channel \times column \times row}$, and is indexed as $\delta_{l, x, y, z}$.
\subsubsection{Weights}
We compute $S=shape(W_{l})$ and we refer to \textbf{Reshape(}$W_{l}$, shape=$(S[0], -1)$\textbf{)} as $W^{f}$.
\begin{equation}{\delta_{l, x, y, z} =  \frac{1}{\beta_{w}}\sum_{r=0}^{\beta_{w}}\left | W^{f}_{(x, r + \beta_{w}*\left \lfloor \frac{(w-1)z + y}{\beta_{w}} \right \rfloor )} \right |}
\label{eq1}
\end{equation}
\begin{equation}{\widetilde{W}_{l} = \delta_{l} \odot Sign(W_{l})}
\label{eq2}
\end{equation}
In \eqref{eq1}, $x \in (0, c),  y \in (0, w), z \in (0, h)$.\\
In most CNNs, filter sizes for convolution rarely exceed $11\times11$. Row-major Hadamard Binarization of the actual filter kernels will not benefit with $\beta_{w} > w_{kernel}$, preventing us from achieving maximum memory saving. Hence we reshape our convolution weight filters as $W^{f} = Reshape(W_{l}, shape=(shape(W_{l}[0], -1))$ for the binarization process. 

\subsubsection{Activations}
\begin{figure} 
\begin{center}
  \includegraphics[width=1\linewidth]{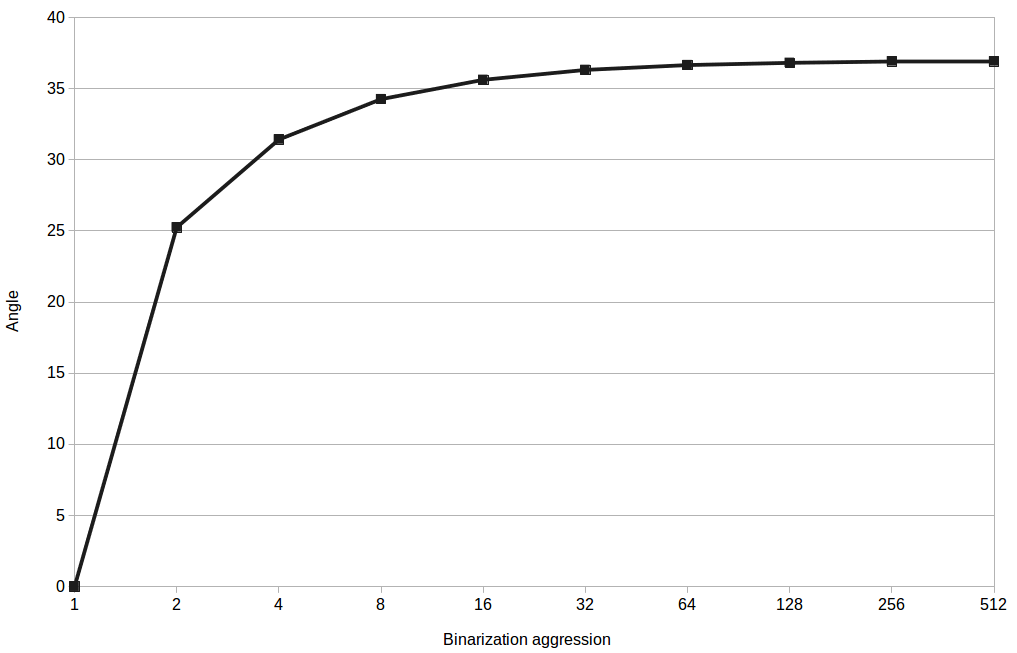}
\end{center}
  \caption{The angle between a random vector and its hadamard binarized vector with respect to the binarization aggression ($\beta$).}
\label{binang}
\end{figure}

\begin{equation}{\delta_{l, x, y, z} =  \frac{1}{\beta_{a}}\sum_{r=0}^{\beta_{a}} \left |
 A_{l, (x, r + \beta_{a}*\left \lfloor \frac{y}{\beta_{a}} \right \rfloor , z)} \right | }
 \label{eq3}
 \end{equation}
\begin{equation}{\widetilde{A}_{l} = \delta_{l} \odot Sign(A_{l})}
\label{eq4}
\end{equation}
In Eqn \eqref{eq3}, $x \in (0, c),  y \in (0, w_{in}), z \in (0, h_{in})$ and $\beta_{a} \leq w_{in}$.
We utilize  the binary weight estimation derived in~\cite{RastegariX1}. The optimal binary estimation of a weight filter is simply $\frac{1}{n}||W||_{l1} \odot Sign(W)$. By segmenting W with windows of size ($1 \times \beta_{w}$) row-wise, a better estimate of a tensor can be developed.
Alexander and Cory~\cite{Alexander17} empirically demonstrate the Angle Preservation Property of XNOR type tensor binarization. Upon studying the angle ($\alpha$) between a random vector (from a standard normal distribution) and its hadamard binarized version for different $\beta$, we observe that as $\beta$ decreased, $\alpha$ diminishes. This is demonstrated in Figure \ref{binang}.

\begin{figure*}
\begin{center}
    \includegraphics[width=1\linewidth]{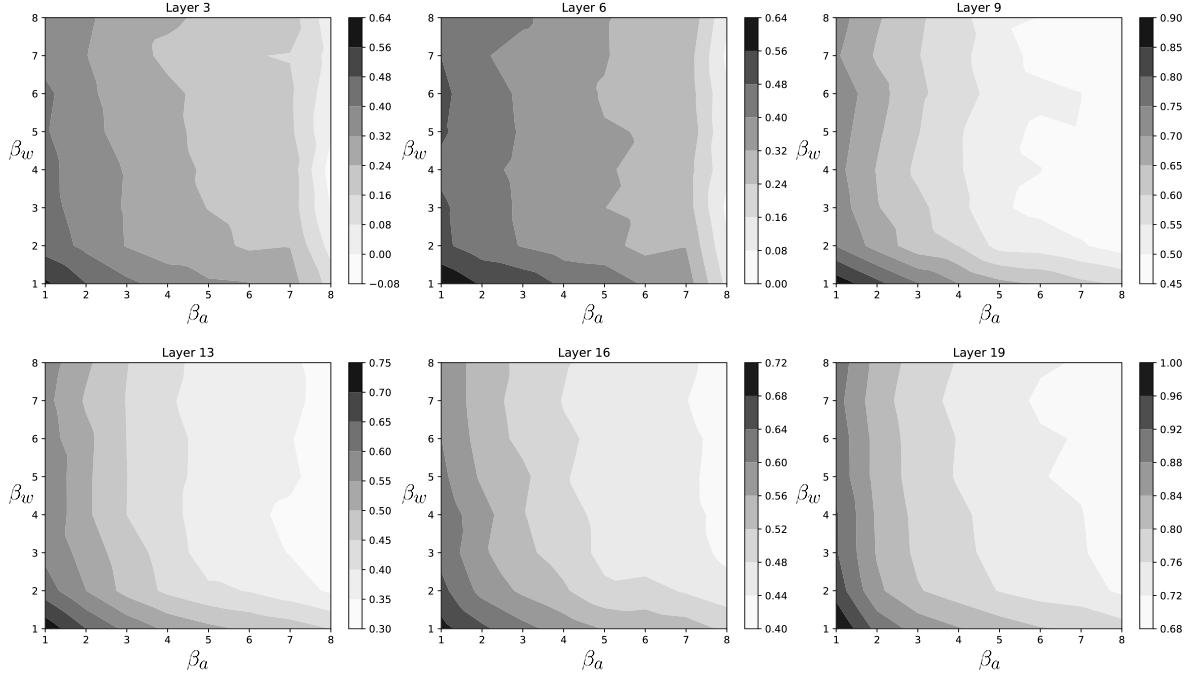}
\end{center}
   \caption{We study the Dot-Product Preservation property for the AlexNet-inspired architecture trained on the CIFAR-10 dataset. The filled contour maps reveal the Pearson's correlation coefficient between the activations in the HadaNet as $\beta_{w}$ and $\beta_{a}$ vary through different layers of the network. $\beta_{w}, \beta_{a} \in \mathbb{Z^{+}}$}
\label{contourmaps}
\end{figure*}

\subsubsection{Propagating gradients}
The derivative of the $Sign$ function is zero almost everywhere. To preserve the gradient information and cancel the gradient when the input ($x$) is too large, we take a straight-through estimator to obtain the derivative of the $Sign$ function \eqref{eq5} ~\cite{CourbariauxB1}. 
\begin{equation}
    {\frac{\partial Sign(x)}{\partial x} = \left\{\begin{matrix}
1_{|x| \leq 1}\\ 
0_{|x| > 1}
\end{matrix}\right.
}
\label{eq5}
\end{equation}
To propagate the gradients through the hadamard binarization function, we assume that the incoming gradient from the $(l+1)^{th}$ layer from a L-layer neural network is $\frac{\partial C}{\partial \widetilde{W}_{l}}$. We require $\frac{\partial C}{\partial W_{l}}$, as given in \eqref{eq6}. 
For simplicity, we take $W \in \mathbb{R}^{n}$. 
Note that $\delta_{l,i}$ references the absolute value of the hadamard binarized weight at the $i^{th}$ index of the $l^{th}$ layer. We refer to $\beta_{w,l}$ as $\beta_{w}$ in Eqn \eqref{eq6}.
\begin{equation} 
    \begin{aligned}
    \frac{\partial C}{\partial W_{l, i}} =  \frac{1}{\beta_{w}} Sign(W_{l, i})\sum_{j=\beta_{w} \left \lfloor \frac{i}{\beta_{w}} \right \rfloor}^{\beta_{w} (\left \lfloor \frac{i}{\beta_{w}} \right \rfloor + 1)} \frac{\partial C}{\partial \widetilde{W_{l, j}}}\left ( Sign(W_{l, j} \right )) \\
    + \hspace{2mm}\delta_{l,i} \frac{\partial C}{\partial \widetilde{W_{l, i}}} \frac{\partial (Sign(W_{l, i}))}{\partial W_{l, i}}
    \end{aligned}
    \label{eq6}
\end{equation}
\subsection{Binarization aggression}
Deciding the values for $\beta_{w,l}$ and $\beta_{a,l}$ in HadaNets is of great importance while making architectural decisions. Extending the Dot-Product preservation study done by Alexander \etal~\cite{Alexander17}, we discover that the activations for HadaNet with varying ($\beta_{w,l}$, $\beta_{a,l}$) and its Full Precision variant for the same architecture are highly correlated. \\
In Figure \ref{contourmaps} we train a neural network (AlexNet-inspired architecture) with $\beta_{a}=8$ and $\beta_{w} = 8$ over the CIFAR-10 dataset. We use this network to find the Pearson's correlation coefficient between the activations for the Network Variant ($\beta_{a}=1$, $\beta_{w} = 1$) and activations for Network Variants with $\beta_{a}$ and $\beta_{w}$ independently varying from 1 to 8. We restrict our $\beta$ to 8 because the activation maps inside this neural network topology have a minimal activation matrix size of $8\times8$. 
It is evident from Figure \ref{contourmaps} that changing $\beta_{w}$ for a given $\beta_{a}$ has little effect on the Pearson's correlation coefficient. Hence we can binarize the weights more aggressively than we binarize the activations. This makes the model more compressible, as the Hadamard binarization of model weights reduce the size of the model $\sim$ $\frac{\beta_{w}X}{\beta_{w} + X}$ times. Here $X$ refers to the bit-width of the available data type to bit-pack the elements of the 'binary' tensor $(\pm 1)$ to. \\
We can also draw from Figure \ref{binang} that the primary benefit of Hadamard Binarization can be observed if $\beta \leq 8$, as the angle ($\alpha$) is seen to saturate  to approximately 37 degees for $\beta > 8$. This could also explain why the Binary-Weight-Network~\cite{RastegariX1} performs significantly better than XNOR-Nets on the ImageNet data-set (A Binary-Weight-Network for the ResNet-18 architecture achieved 60.8\% top-1 accuracy, whereas an XNOR-Network had a top-1 accuracy of 51.2\%). \\
We also observe that for our AlexNet-inspired architecture, changing the $\beta_{w}$ for the $3^{rd}$ and $4^{th}$ convolutional layers causes negligible difference ($\pm 0.5\%$) in CIFAR-10 accuracy. We tested this for $\beta_{w, l\in\{3, 4\}} = \{16, 32, 64\}$. We observe that we could binarize the $2^{nd}$ convolution with $\beta_{w} = 32$ for Le-Net architecture for the MNIST data-set with no degradation in accuracy ($\pm 0.1\%$). With the above, we reason that we can yield even greater memory savings for very large scale image recognition models by aggressive layer-specific binarization with little or no loss in accuracy. We discuss the different $\beta$ configurations we tested over standard data-sets in Section 5 of this paper. 

\begin{figure} 
\begin{center}
  \includegraphics[width=1\linewidth]{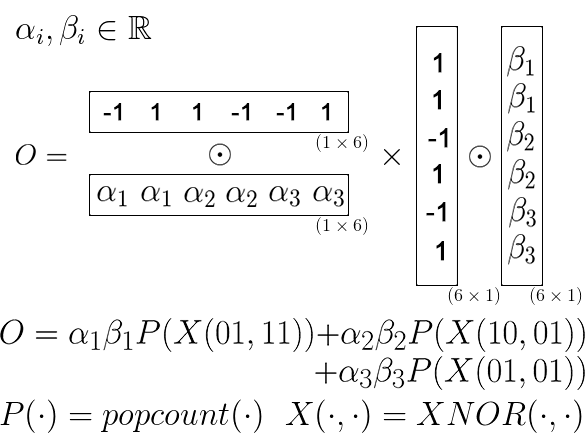}
\end{center}
  \caption{The figure above demonstrates the hadamard multiplication (xHBNN) of two vectors. Here $\beta_{a} = \beta_{w} = 2$}    
\label{Hadamultiplication}
\end{figure}

\section{Efficiency analysis}
A majority of current hardware implementations are variants of von-Neumann machines~\cite{Agrawal2018XcelRAMAB}. In such machines, the memory and computation blocks are separate. This is a bottleneck as CNNs require constant data transfer between the memory and computational blocks. A single 32 bit DRAM memory access takes 640pJ in 45nm CMOS technology, whereas a floating point add consumes 0.9pJ~\cite{DeepCompress}. In this regard, binary neural networks are very memory friendly. It is possible, and $128\times$ more power efficient to store some models in the SRAM cache of the device, where a 32 bit SRAM memory access takes 5pJ.
Courbariaux \etal~\cite{CourbariauxB1} states that due to binary convolution kernel repetitions, the dedicated hardware time complexity is reduced by 60\%. While this is true, kernel repetitions restrict the model accuracy, in the same way increasing kernel depth in a Binary Weight Neural network might be futile because of the limited number of filter kernels possible ($2^{n\times m}$) for a filter of size $(n \times m)$. HadaNets are less susceptible to learning redundant filters as we have greater degree of freedom (A full precision tensor instead of a scalar tied to a binary tensor).\\
The energy spent to move data from SRAM to register or register to register increases with the length of the word size. In our networks, the smaller the $\beta$, the higher the network accuracy. It can therefore be beneficial to bit-pack the binary matrix resulting from Hadamard Binarization to smaller data-types (8-bits). As an example, from Figure \ref{Hadamultiplication} we can see that it is possible to bit-pack the row vector $'{-1}\ 1\ 1\ {-1}\ {-1}\ 1'$ as $011001$ to a 6-bit data-type. To compute $O$, we can get to the relevant part of the binary matrix by doing bit-shifts. Packing the binary matrices to a smaller data-type not only reduces the energy consumption in moving the data, but also reduces the amount of bit-shifts required to obtain the relevant part of the binary matrix for bit-manipulations. \\
A quadratic increase in parameters for ABC-Nets~\cite{LinABC} makes maintaining full-precision parameters during training costly. HadaNets side step the issue of a quadratic rise in parameter count, and successfully reduce the train time and cost. The issue of creating too many activation/weight bases with high correlation in the ABC Nets is dealt with by $l_{2}$ regularization. This inherent problem with ABC-Nets does not exist in HadaNets. \\ 
In all our experiments and tests reported, we do row-wise binarization for all $\beta_{w}$ and $\beta_{a}$ variants. We assume that all tensors are stored in a row-major order. We avoid segmenting tensors into tensor blocks as this will increase the memory access times significantly irrespective of row/column major storage. 
\begin{table}
\begin{center}
\begin{tabular}{|l|c|}
\hline
Network & Memory Saving \\
\hline\hline
AlexNet & 6.51$\times$ \\
ResNet-18 & 7.43$\times$ \\
\hline
\end{tabular}
\end{center}
\caption{Reduction in model size for $\beta_{w} = 16$ with respect to the full precision model.}
\label{table:MemoryReducing}
\end{table}
\subsection{Memory}
As discussed earlier, the memory required for a HadaNet model is $\sim$ $\frac{\beta_{w}X}{\beta_{w} + X}$ times lesser than its full precision counterpart. For a FP32 model, we shall assume that the binary elements are bit-packed to an unsigned long int, making $X = 32$. 
Memory accesses cost significantly more energy than arithmetic operations in a neural network. Table \ref{table:MemoryReducing} details how aggressive the memory saving is for ResNet-18 and AlexNet. For both models, we shall take $\beta_{w} = 16$, and keep $\beta_{w, 1} = \beta_{w, L} = 1$.
Rastegari \etal~\cite{RastegariX1} found that the scaling factor for weights is more important than the scaling factor for the activations. We suspect that this observation for an XNOR-Net is valid because the weight tensors compensate for the absence of the activation scaling factor at train-time. 
A Binary-Weight-Network, where no binarization is done on the activations outperform XNOR networks. We also found that $\beta_{a}$ played a more important role in accuracy than $\beta_{w}$, with $\beta_{a} = 2$ and $\beta_{w} = 16$ $(\beta_{w} \geq \beta_{a})$ outperforming models where $\beta_{w} \leq \beta_{a}$ on the CIFAR-10 data-set. We also conducted tests on the ResNet-18 keeping $\beta_{a} = 1$ and $\beta_{w} = 4$ and observe that the network outperformed the Binary-Weight-Network by 2.5\% in top-1 accuracy. These observations pave way for greater binarization aggression for weights, which further reduce the memory of trained networks. 
It has been observed that at train-time, storing the activation maps for mini-batches creates a larger memory footprint than the weights do. This has not been effectively resolved by HadaNets. 
\subsection{CMMA vs xHBNN}
\begin{figure}
\begin{center}
  \includegraphics[width=1\linewidth]{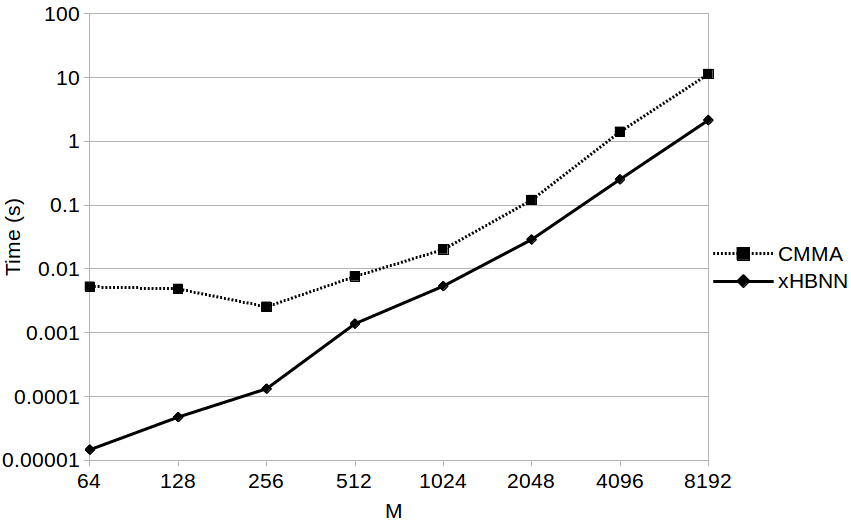}
\end{center}
  \caption{Benchmarking CMMA (Classical matrix multiply algorithm) with the Hadamard Binary Matrix Multiplication (xHBNN) kernel. The M denotes the size of a square matrix $(M \times M)$. In this figure, $\beta = 16$.}
\label{xhbnnmark}
\end{figure}

Figure \ref{xhbnnmark} benchmarks the Classical Matrix Multiply Algorithm with our Hadamard Binary Matrix Multiplication kernel. We keep the $\beta = 16$ while multiplying the matrices. We used the Intel Xeon Gold 6128 processor clocked at 3.40 GHz, a 19.25 MB cache, and 24 cores with two-way Intel Hyper-Threading Technology in our benchmark. We observe an approximate 10$\times$ speed up when using the xHBNN kernel for matrix multiplication.\\
As discussed in~\cite{CourbariauxB1}, we can concatenate groups of $\beta$ binary variables into available data types, and replace $\beta$ multiplication and $\beta - 1$ sum operations with two multiplication and simple bit-wise operations.\\
To compute the product of two vectors $W$ and $A \in \mathbb{R}^{K}$ where $W_{f}$, $A_{f}$ $\in \mathbb{R}^{\left \lceil \frac{K}{\beta} \right \rceil}$ and $W_{b}, A_{b} \in \{-1, +1\}^{\left \lceil \frac{K}{\beta} \right \rceil}$, the follow operations shall be necessary:
\begin{equation}
    \begin{aligned}
    {
    O = \sum_{j=0}^{\left \lceil \frac{K}{\beta} \right \rceil} W_{f, j} \times A_{f, j} \times popcount(xnor(W_{b, j}, A_{b, j}))
    }
    \end{aligned}
    \label{eq7}
\end{equation}
It is important to note that Eqn \eqref{eq7} refers to $W_{b}, W_{f}, A_{b}, A_{f}$ after the 'paired' full precision and binary vector approximations of $W$ and $A$ have been compressed. The binary vector is bit-packed to available data types, and the full precision vector can be compressed $\beta$ times. For brevity, we assume that there is a data-type with bit-width $ = \beta$ available, thus compressing the binary vector to a size of $\left \lceil \frac{K}{\beta} \right \rceil$. This method of multiplication has been demonstrated in Figure \ref{Hadamultiplication}.\\
It is evident that if $\beta_{a} \leq \beta_{w}$, the inference speed of the network will be decided by $\beta_{a}$. This places a potential bottleneck in computation time. We reason that since a parallel reduction approach computes the summation or multiplication of N-1 numbers in $log_{2} (N)$ steps; changing $\beta_{a}$ from 4 to 16 would not give significant speedups and will cause accuracy deterioration. For a 45nm CMOS technology a 64-bit memory access from an 8K cache takes about 10pJ~\cite{EnergyHorowitz}, whereas a 32-bit FMUL operation takes 3.7pJ. Typical SRAM access latency is around 2-3 ns, whereas DRAM access latency is approximately 20-35 ns. We reason that it is a better pursuit to reduce the size of the model instead of the number of arithmetic operations. The primary speed up will be realized by placing model weights in the cache. As the batch-size increases, the memory footprint of activations increase. It was found in~\cite{Mishra17} that about $96.5\%$ of the memory footprint for the ResNet-101 at inference with a batch-size of 1 was due to weights. Thus the memory saving during inference for most real-time scenarios will primarily come from keeping $\beta_{w}$ low.
\begin{table*}[t]
  \begin{center}
  \begin{tabular}{|c|c|c|c|c|}
  \hline
     \multicolumn{2}{|c|}{Data-set} & \multicolumn{1}{|c|}{MNIST} &
     \multicolumn{2}{|c|}{CIFAR-10} \\
    \hline \hline
         \multicolumn{1}{|c|}{Network Variant} & \multicolumn{1}{|c|}{Memory} & \multicolumn{1}{|c|}{LeNet} & \multicolumn{1}{|c|}{NIN} & \multicolumn{1}{|c|}{AlexNet-inspired}  \\
            \hline
             Full-Precision & $1\times$ & 99.39\% & 89.59\% & 89.36\% \\
             HBWN ($\beta_{w}=4; \beta_{a}=1$) & $\sim.28\times$ & 99.41\% & 89.33\% & 89.24\% \\
             HadaNet ($\beta_{w}=4; \beta_{a}=4$)  & $\sim.28\times$ & 99.43\% & 87.33\% & 88.64\% \\
             ABC Net ($b_{w}=5; b_{a}=1$) & $\sim.16\times$ & 99.41\% & - & 88.69\% \\
             HBWN ($\beta_{w}=8; \beta_{a}= 1$) & $\sim.16\times$ & 99.41\% & 89.11\% & 89.04\% \\
             HadaNet ($\beta_{w}=16; \beta_{a}=2$) & $\sim.09\times$ & 99.40\% & 88.74\%
             & 89.02\% \\ 
             XNOR-Net & $\sim.03\times$ & 99.23\% & 86.28\% & 88.60\% \\
            \hline
  \end{tabular}
  \end{center}
  \caption{Classification test accuracy of CNNs trained on MNIST and CIFAR-10 with different network topologies.}
  \label{MNISTCIFARaccuracy}
\end{table*}
\begin{table*}
  \begin{center}
  \begin{tabular}{|c|c|c|c|c|c|c|}
  \hline
     \multicolumn{7}{|c|}{ImageNet} \\
    \hline \hline
         \multicolumn{1}{|c|}{Network Topology} &
         \multicolumn{3}{|c|}{AlexNet} & \multicolumn{3}{|c|}{ResNet-18}  \\
            \hline \hline
            Network variant & Top-1 & Top-5 & Memory & Top-1 & Top-5 & Memory \\
             \hline
              Full-Precision & 56.6\% & 80.2\% & $1\times$ & 69.3\% & 89.2\% & $1\times$ \\
              HBWN ($\beta_{w} = 4; \beta_{a} = 1$) & 56.7\% & 80.1\% & $\sim.32\times$ & 62.3\% & 84.4\% & $\sim.31\times$\\
              HadaNet ($\beta_{w} = 4; \beta_{a} = 4$) & 46.3\% & 71.2\% & $\sim.32\times$ & 53.3\% & 77.3\% & $\sim.31\times$\\
              ABC Net ($b_{w}=5; b_{a}=1$) & - & - & $\sim.19\times$ & 54.1\% & 78.1\% & $\sim.17\times$\\
              ABC Net ($b_{w}=5; b_{a}=3$) & - & - & $\sim.19\times$ & 62.5\% & 84.2\% & $\sim.17\times$\\
              HadaNet ($\beta_{w} = 16; \beta_{a} = 2$) & 47.3\% & 73.3\% & $\sim.13\times$ & 53.8\% & 77.2\% & $\sim.12\times$\\
              ABC Net ($b_{w}=3; b_{a}=1$) & - & - & $\sim.12\times$ & 49.1\% & 73.8\% & $\sim.10\times$\\
              BWN & 56.8\% & 79.4\% & $\sim.05\times$ & 60.8\% & 83.0\% & $\sim.05\times$ \\
              XNOR-Net & 44.2\% & 69.2\% & $\sim.05\times$ & 51.2\% & 73.2\% & $\sim.05\times$ \\
              BNN &  27.9\% & 50.4\% & $\sim.04\times$ & 42.2\% & 67.1\% & $\sim.04\times$\\
              
            \hline
  \end{tabular}
  \end{center}
  \caption{Classification test accuracy of CNNs trained on the ImageNet dataset with different network topologies. Note that the memory column only estimates model sizes, and does not describe the run-time memory overhead. }
  \label{ImageNetaccuracy}
\end{table*}
\section{Experiments}
In this section we evaluate the performance of HadaNet variants with respect to full precision neural networks, XNOR-Nets~\cite{RastegariX1} and ABC Nets~\cite{LinABC}. We train our models on the MNIST~\cite{lecun-mnisthandwrittendigit-2010}, CIFAR-10~\cite{CIFAR10} and the ImageNet (ILSVRC2012)~\cite{ILSVRC15} datasets. XNOR-Net~\cite{RastegariX1} is a neural network where both the filters and input to convolution layers are binary, with a full precision scalar scaling factor for the activations and weight tensors. HadaNet with $\beta_{a,l} = nElement(A_{l})$ and $\beta_{w, l} = nElement(W_{l})$ (where $nElement$ stands for the number of elements in the weight or activation tensor) is essentially the XNOR-Net. ABC Nets approximate full-precision weights as a linear combination of multiple binary weight bases. We report results for ABC Nets with a 5-weight base, 1-activation base. HadaNet with $\beta_{a} = 2; \beta_{w}=16$ has a comparable memory footprint. In all our experiments, the first and the last layer have $\beta_{w} = \beta_{a} = 1$. In all trials where $\beta_{w,l}, \beta_{a,l} \neq 1$, we use the following placement of layers: Batch Normalization - Binarization - Convolution - Activation - Pooling, as suggested by Rastegari \etal~\cite{RastegariX1}, the Convolution - Batch Normalization - Binarization - Pooling placement of layers reduced the top-1 accuracy of the XNOR-Network by approximately 14\%~\cite{RastegariX1}.
\subsection{MNIST} 
 We train on the MNIST~\cite{lecun-mnisthandwrittendigit-2010} dataset for 60 epochs and decay the learning rate by 0.1 every 15 epochs with an initial learning rate of 0.005. We minimize the Cross-Entropy Loss with the Adam optimizer. We train the LeNet topology and use Batch Normalization with a minibatch of size 128. No data-augmentation is done.
\subsection{CIFAR-10}
We train two network topologies on the CIFAR-10~\cite{CIFAR10} data-set. The first is the Network-In-Network architecture~\cite{NINnet}. The second is an AlexNet-inspired network with the following architecture:\\
$(2\times 128C3) - MP2 - (2\times256C3)-MP2-(2\times512C3)-MP2-(2\times1024FC)-10FC$\\
C3 is a batch normalized 3$\times$3 ReLU convolution layer. $\beta_{a}$ $=$ $\beta_{w} = 1$ for the first and the last layer. We minimize the Cross-Entropy Loss with the Adam optimizer and an exponentially decaying learning rate. We train our neural network with a batch size of 128 for 60 epochs. We trained the ABC Net on the CIFAR-10 dataset as it was not reported in their paper. 
We do not use any data-augmentation techniques. We simply Normalize the data-set.\\
\subsection{ImageNet}
ImageNet (ILSVRC2012) is a benchmark image classification dataset~\cite{ILSVRC15} which consists of ~1.2 million training images. There are 1000 categories.\\
\textbf{AlexNet}\\
The AlexNet network has 5 convolutional layers and 2 fully-connected layers. This network has 61 million parameters. We resize our ImageNet dataset to a size of 256$\times$256 and take random-crops of size 227$\times$227. We augment our dataset with random horizontal flips. We minimize the Cross-Entropy Loss with the ADAM optimizer and set our initial learning rate as 0.001. We train our network for 32 epochs, decaying the learning rate by 0.1 after every 8 epochs. \\
\textbf{ResNet-18}\\
We evaluate the ResNet-18~\cite{behemothHe_2016_CVPR} as done in~\cite{RastegariX1}. We train for 60 epochs with a batch size of 256 and initial learning rate of 0.01, we decay our LR by 0.1 at epoch 30 and 40. We augment our dataset as done in the AlexNet training procedure but crop the input to 224$\times$224. 
\section{Discussion}
In most of our trials, we keep $\beta_{a} \leq \beta_{w}$. Increasing $\beta_{a}$ significantly degrades the network accuracy. The primary issue with quantization of activations is that we need to introduce an approximation for the non-differentiable $Sign$ operator. The derivative is zero almost everywhere for the $Sign$ function. We use a linear approximation for the $Sign$ function as detailed in Eqn \eqref{eq5}. This does not solve the gradient mismatch problem. We keep $\beta_{a} \in [1, 8]$ in our experiments. The benefits of Hadamard Binarization become less pronounced for $\beta_{a} > 4$, but is relatively flexible to changes in $\beta_{w}$. \\
Drawing from the findings in ABC Nets~\cite{LinABC}, we tested Average Pooling instead of Max Pooling over the CIFAR-10 dataset. This was because in their experiments, max-pooling returned a tensor with most elements equal to $+1$. We also used the PReLU activation function and tested the AlexNet-inspired network on the CIFAR-10 dataset. The PReLU activation function is initialized with $a_{i} = 0.25$. These tests did not yield any noticeable gain in performance for HadaNets.\\
ABC Nets~\cite{LinABC} use pre-trained models to convert an optimization problem to a linear regression problem. The activation maps generated by ABC Nets occupy a larger memory footprint than HadaNets during inference. Utilizing 5 binary activation maps to represent a full precision activation map along with 5 binary weight maps to approximate a full precision weight map would generate a memory footprint close to that of its full-precision counterpart at inference. Unlike ABC Nets, HadaNets are networks that are trained from scratch. If a variant of ABC Net is trained from scratch, the memory overhead generated would be far greater than HadaNets. Training a full-precision model and converting the model to ABC Nets would also require more energy and time than a HadaNet. \\
Our current training methods utilize different forward and backward approximations, which gives rise to the gradient mismatch problem. Drawing from~\cite{Leng2018ExtremelyLB}, it is possible to formulate our HadaNets training as a discretely constrained optimization problem and decouple the continuous parameters from the discrete constraints. Leng \etal~\cite{Leng2018ExtremelyLB} solve this problem using extra-gradient and iterative quantization algorithms. While HadaNets are compatible with this framework, implementing this was beyond the scope of this paper.
\section{Conclusion}
We introduce HadaNets, which utilize a new weight and activation binarization scheme. This method of binarization does not increase the parameter count of the neural network, and works with the hyper-parameters $\beta_{w,l}$ and $\beta_{a,l}$ that can be tuned to match hardware with a range of compute capabilities and memory constraints. We empirically justify the importance of $\beta_{a}$ in accuracy and reinforce our claim by demonstrating the dot product preservation property and the Angle Preservation Property for HadaNets. We train several HadaNet variants and reported our results over the MNIST, CIFAR-10 and ImageNet dataset. We outperform the XNOR-Net and ABC Net over the MNIST and CIFAR-10 dataset. Our AlexNet network variant ($\beta_{w} = 16; \beta_{a} = 2$) out-performs the XNOR Net and give accuracy that is at par with full precision neural networks. Our Hadamard-Weight-Binary-Network outperforms Binary-Weight-Networks by 1.5\% in top-1 accuracy for ResNet-18. We also demonstrate a Hadamard Binary Matrix Multiplication CPU kernel (xHBNN) which delivered a $10\times$ speed up over a similarly optimized Classical Matrix Multiplication CPU kernel (CMMA). Developing highly efficient quantized neural networks requires novel solutions to quantize weights, activations and gradients. Future work could involve using pre-trained networks to initialize HadaNets, testing more network variants on state-of-the-art network topologies. Gradient quantization methods for more efficient distributed training of neural networks should also be studied in greater detail. 



{\small
\bibliographystyle{ieee}
\bibliography{egbib}
}

\end{document}